\documentclass[preprint,12pt]{elsarticle}

\usepackage{graphicx}
\usepackage{amssymb}

\usepackage{amsmath, amsthm, amsfonts,amssymb, bm}
\usepackage{svg}
\usepackage{subfig}
\usepackage{float}
\usepackage{url}
\usepackage{enumerate}
\usepackage{color}   
\usepackage{hyperref}
\hypersetup{
    colorlinks=true, 
    linktoc=all,     
    linkcolor=blue,  
}
\usepackage[ruled,vlined,linesnumbered]{algorithm2e}
\usepackage{bm}
\usepackage{tikz}
\usepackage{mathdots}
\usepackage{yhmath}
\usepackage{cancel}
\usepackage{color}
\usepackage{siunitx}
\usepackage{array}
\usepackage{multirow}
\usepackage{amssymb}
\usepackage{gensymb}
\usepackage{tabularx}
\usepackage{booktabs}
\usepackage{graphicx}
\usepackage[demo]{rotating}
\usetikzlibrary{fadings}
\usetikzlibrary{patterns}
\usetikzlibrary{shadows.blur}

\usepackage{xspace} 

\usepackage[export]{adjustbox}
\usepackage{makecell}
\usepackage{multirow}

\newcommand{\model}{\texttt{DANLIP}\xspace}

\allowdisplaybreaks

\begin{document}

\begin{frontmatter}


\title{DANLIP: Deep Autoregressive Networks for Locally Interpretable Probabilistic Forecasting}

\author[1]{Ozan Ozyegen}
\author[2]{Juyoung Wang}
\author[1]{Mucahit Cevik}
\address[1]{Department of Mechanical and Industrial Engineering, Toronto Metropolitan University, Canada}
\address[2]{Department of Mechanical and Industrial Engineering, University of Toronto, Canada}

\begin{abstract}
Despite the high performance of neural network-based time series forecasting methods, the inherent challenge in explaining their predictions has limited their applicability in certain application areas. 
Due to the difficulty in identifying causal relationships between the input and output of such black-box methods, they rarely have been adopted in domains such as legal and medical fields in which the reliability and interpretability of the results can be essential. 
In this paper, we propose \model, a novel deep learning-based probabilistic time series forecasting architecture that is intrinsically interpretable. 
We conduct experiments with multiple datasets and performance metrics and empirically show that our model is not only interpretable but also provides comparable performance to state-of-the-art probabilistic time series forecasting methods. 
Furthermore, we demonstrate that interpreting the parameters of the stochastic processes of interest can provide useful insights into several application areas. 
\end{abstract}


\end{frontmatter}


\section{Introduction}

Interpretable machine learning, a subfield of statistical learning, has received a lot of attention in the last few years. 
Despite the significant achievements of statistical learning techniques, such methods are not fully adopted in some decision-critical settings such as those frequently seen in legal and medical domains~\citep{melis2018towards}. 
It is primarily because, in such cases, decision-makers are typically required to answer the question \textit{``Why did the machine learning model make such decisions?''}, as described by \citet{molnar2020interpretable}. 

According to \citet{doshi2017towards}, interpretability in the context of machine learning is defined as ``\textit{the ability to explain or to present in understandable terms to a human}''. 
Under such a criterion, naive neural network-based learning methodologies fail to achieve interpretability due to their black-box nature. 
Specifically, the neuron activation status of the network is likely to fail to provide \textit{human-level understanding} of how the neural network makes the prediction.

Linear regression is frequently used as a traditional forecasting algorithm and it can be considered a typical example of interpretable statistical learning methodologies. 
This method allows users to express the predictions as a linear combination of covariates.  
That is, the model assumes the response variable expressed as
\begin{align}
\label{eqn:lr}
    y = \sum_{f \in \mathcal{F}} \gamma_f x_f + \varepsilon
\end{align}
where $y$ is target, $\mathcal{F}$ is set of features, $\gamma_f \in \mathbb{R}$ is linear model coefficient for all $f \in \mathcal{F}$, and $\varepsilon$ is a Gaussian noise.
Standardized linear regression is regarded as an interpretable forecasting method, as each linear combination coefficient $\gamma_f$ for feature $f \in \mathcal{F}$ can be taken as a degree of importance of that feature.

While linear regression has been extensively used, the method is not always applicable in practice. 
Its functional form restriction inherently limits the expressive abilities of the model, as there exist many complex functions that cannot be expressed as a linear combination of features. 
In this regard, to achieve better performance, the use of complex learning methods is usually inevitable.
However, such methodologies, e.g., deep neural networks, are not necessarily interpretable.

Ensuring the interpretability of the model is extremely important in many disciplines~\citep{melis2018towards}. 
Even in some domains where interpretability is not deemed to be as important, interpretable models can greatly help decision-makers to make more informed decisions. 
For instance, in the case of a large retail company, understanding the key factors that highly affect the sales of flagship products might be of great interest in designing better production, inventory management, and marketing policies. 

There exist two main branches in the realm of interpretable machine learning~\citep{molnar2020interpretable}.
\textit{Post-hoc} methods consist of applying techniques that analyze the model after training, e.g., LIME~\citep{ribeiro2016should} and SHAP~\citep{lundberg2017unified}.
On the other hand, \textit{intrinsic} methods involve building inherently interpretable models. 
This is commonly achieved by restricting the complexity of the machine learning model itself, e.g., RETAIN~\citep{choi2016retain}.

In general, post-hoc methods such as SHAP are slow and not efficient for large-scale machine learning problems. 
These methods usually require a large number of experiments to calculate interpretability scores.
Indeed, the runtime complexity of the SHAP method is exponential in the number of features in the dataset~\citep{molnar2020interpretable}. 
In contrast, intrinsically interpretable methodologies generally work fast. 
It has been shown that the performance of proposed methods can be comparable to the non-interpretable methods~\citep{choi2016retain}. 
Accordingly, we concentrate on developing an intrinsically interpretable time series forecasting method using neural networks. 
As intrinsic methods typically have a faster training time, this framework can be more suitable to develop prediction algorithms over a large amount of training data.

Only a few studies have explored interpretable deep neural network-based time series forecasting methods. 
Among these works, Temporal Fusion Transformer (TFT)~\cite{lim2021temporal} is highly related to our work, as the architecture also achieves interpretability, while performing time series forecasts. 
However, the TFT is fundamentally different from our work as it only computes \textit{importance} scores rather than \textit{contribution} scores of features. 
Unlike importance scores, contribution scores allow users to quantify both the positive and negative impact of features on the prediction. 
We summarize the contributions of our study as follows:
\begin{itemize}\setlength\itemsep{0.13em}
    \item We present DANLIP, a deep learning-based interpretable parametric probabilistic time series forecasting model. 
    
    \item We show our model can achieve comparable performance to those of state-of-the-art probabilistic time series forecasting methods while maintaining high interpretability. 
    We conduct an extensive numerical study with three real-world datasets to show the performance of our proposed method.
    
    \item We empirically show how interpreting the predictions of the model using the parameters of stochastic processes that are fitted to the model can be used to provide important practical insights.
\end{itemize}

The rest of this paper is structured as follows. 
In Section~\ref{sec:bw}, we briefly review the relevant literature with a specific focus on RETAIN~\citep{choi2016retain} and DeepAR~\citep{salinas2020deepar} models as two highly relevant approaches for probabilistic forecasting. 
In Section~\ref{sec:POLAR}, we present our interpretable time series forecasting model. 
Throughout the section, we describe the model architecture and show how the prediction results can be interpreted in terms of both mean and variance. 
In Section~\ref{sec:exp}, we show the performance of our model empirically using multiple datasets evaluated over various performance metrics. 
Moreover, we discuss how interpreting both the mean and variance of the probabilistic forecasts can provide useful insights using real-world datasets. 
Finally, in Section~\ref{sec:conclusions}, we summarize our findings and discuss future work.

\section{Background}\label{sec:bw}
In this section, we first briefly review the most relevant studies from the literature and summarize time series forecasting basics.
Then, we review the studies by \citet{salinas2020deepar} and \citet{choi2016retain} in detail, which provide the preliminaries for our own methodology.

\subsection{Related works}
There is a vast literature on time series forecasting.
Earlier studies in the field mainly focused on statistical models such as Moving Average (MA), Autoregressive Moving Average (ARMA), and Autoregressive Integrated Moving Average (ARIMA)~\citep{Box2015}.
Decision trees and their ensembles, namely gradient boosted trees and random forests were also frequently used for time series forecasting as they possess interpretable structures and have the ability to capture nonlinear relations between various features in the dataset~\citep{galicia2019multi, makridakis2020m5}.

While the majority of the previous studies focus on point forecasting, several recent studies proposed methodologies that are designed to perform high-quality probabilistic forecasts.
A probabilistic forecast typically refers to the confidence interval around the point forecasts and is specified by the lower and upper limits. 
Standard methods (e.g., ARIMA and exponential smoothing) can generate probabilistic forecasts through closed-form expressions for the target predictive distribution or via simulations~\citep{Box2015}. 

As more and more data has become available over the recent years, many studies have focused on deep learning-based approaches for time series forecasting.
For instance, recent studies by \citet{rangapuram2018deep} and \citet{salinas2020deepar} proposed deep learning models for probabilistic forecasting that can directly predict the parameters (e.g., mean and variance) of the probability distribution that specifies the probabilistic forecast. 
These approaches show substantial performance improvements over standard approaches for datasets which consist of a large number of time series (e.g., in the order of hundreds or thousands).
Despite the success of these deep learning methods for forecasting, they came with an important caveat of having highly complex, black-box architectures, which lack interpretability and explainability. 
We refer the reader to recent review articles by \citet{parmezan2019evaluation} for a detailed overview of statistical and machine learning models for time series forecasting.
In addition, recent forecasting competitions provide valuable insights on best-performing time series forecasting methods~\citep{bojer2021kaggle, makridakis2020m4, makridakis2020m5}.

Explainable artificial intelligence also gained significant attention in recent years.
Post-hoc interpretability methods have been used to interpret the decisions of time series models. 
\citet{felix2019shap} used SHAP to interpret time series classifiers, whereas \citet{ozyegen2020evaluation} evaluated three post-hoc interpretability methods, including SHAP, to interpret the time series forecasting models. 
On the other hand, many time series forecasting methods take into account interpretability considerations in model development~\citep{choi2016retain, ilic2021explainable}.
While intrinsically interpretable methods usually trade off the performance to provide more interpretability, the level of this trade-off varies according to the model and the prediction task.
It has been shown that some intrinsically interpretable models achieve similar performance levels to those of black-box nature~\citep{li2021learning, lim2021temporal}. 
Our study fundamentally differs from prior work on intepretable time series forecasting, as the proposed architecture is able to discern which features positively and negatively affect the prediction outcome. 

\subsection{Preliminaries}
In our analysis, we mainly rely on the standard neural network methodology. For the sake of convenience, we additionally introduce the following notations:
\begin{itemize}
    \item $[N] = \{1,2, \hdots, N\}$ for any $N \in \mathbb{N}$.
    \item Given $A \in \mathbb{R}^{m \times n}$, $A[i,:]$ denotes the $i^{\text{th}}$ row of matrix $A$, and $A[:, j]$ denotes the $j^{\text{th}}$ column of matrix $A$. For one- and multi- dimensional tensors, the notations are analogously extended.
    \item Given $A \in \mathbb{R}^{m \times n}$, $A[i_1:i_2,:]$ denotes $(A_{ij})_{i \in \{i_1, \hdots, i_2 - 1\}, j \in [n]}$. The definitions for $A[:,j_1:j_2]$ and $A[i_1:i_2,j_1:j_2]$ are analogous. For one- and multi- dimensional tensors, the notations are analogously extended.
    \item $\mathbb{I}_{\text{condition}}$ is an indicator function which takes value 1 if the condition provided as the subscript is satisfied, and 0 otherwise.
    \item For any $a \in \mathbb{R}^A$ and $b \in \mathbb{R}^B$,  $[a,b]$ refers to concatenation of two vectors, i.e., $[a,b] = [a_1, \cdots, a_A, b_1, \cdots, b_B]$.
\end{itemize}

Following the self-explaining neural network model proposed by \citet{melis2018towards}, which is analogous to the RETAIN architecture of \citet{choi2016retain}, we ensure the interpretability of our forecasting method by restricting the form of the output of the neural network as
\begin{align}
    \label{eqn:nlc}
    y = \sum_{f \in \mathcal{F}} \gamma^{\Theta}(x)_f x_f.
\end{align}
Equation~\eqref{eqn:nlc} is highly similar to Equation~\eqref{eqn:lr}, 
however, different from Equation~\eqref{eqn:lr}, the term that is multiplied with $x_f$ is now a function of input, $\gamma^{\Theta}: D \to \mathbb{R}^{|\mathcal{F}|}$, where $D$ represents the domain of the input data. 
We characterize the function $\gamma^{\Theta}(\cdot)$ with $\Theta$, which denotes the set of neural network parameters.

\subsection{Review of DeepAR architecture}
\begin{figure}[!ht]
    \centering
\includegraphics[width=0.9285\textwidth]{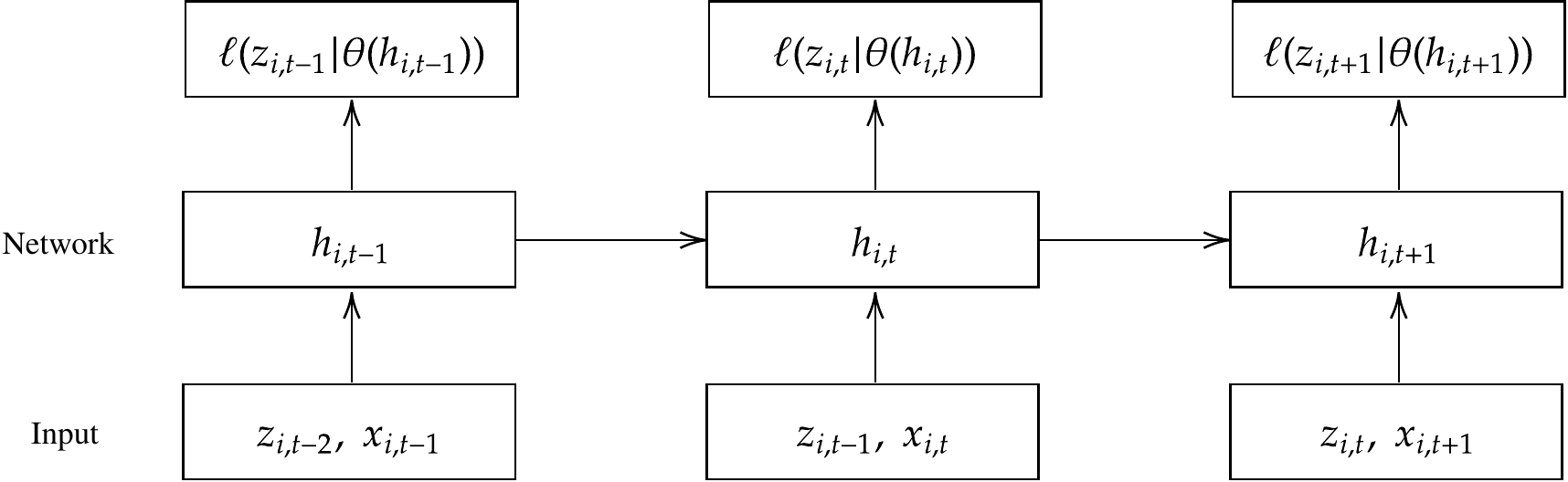}
\caption{Simplified architecture of DeepAR}
\label{fig:DeepAR}
\end{figure}

We consider DeepAR~\citep{salinas2020deepar}, a state-of-the-art probabilistic parametric time series forecasting method, as a baseline in our numerical study.
DeepAR is an RNN-based parametric probabilistic time series forecasting architecture, which seeks to solve the log-likelihood maximization problem. 
Figure~\ref{fig:DeepAR} provides an intuitive visual summary of the DeepAR architecture.

\subsection{Review of RETAIN architecture}
\begin{figure}[!ht]
\centering
\includegraphics[width=0.6585\textwidth]{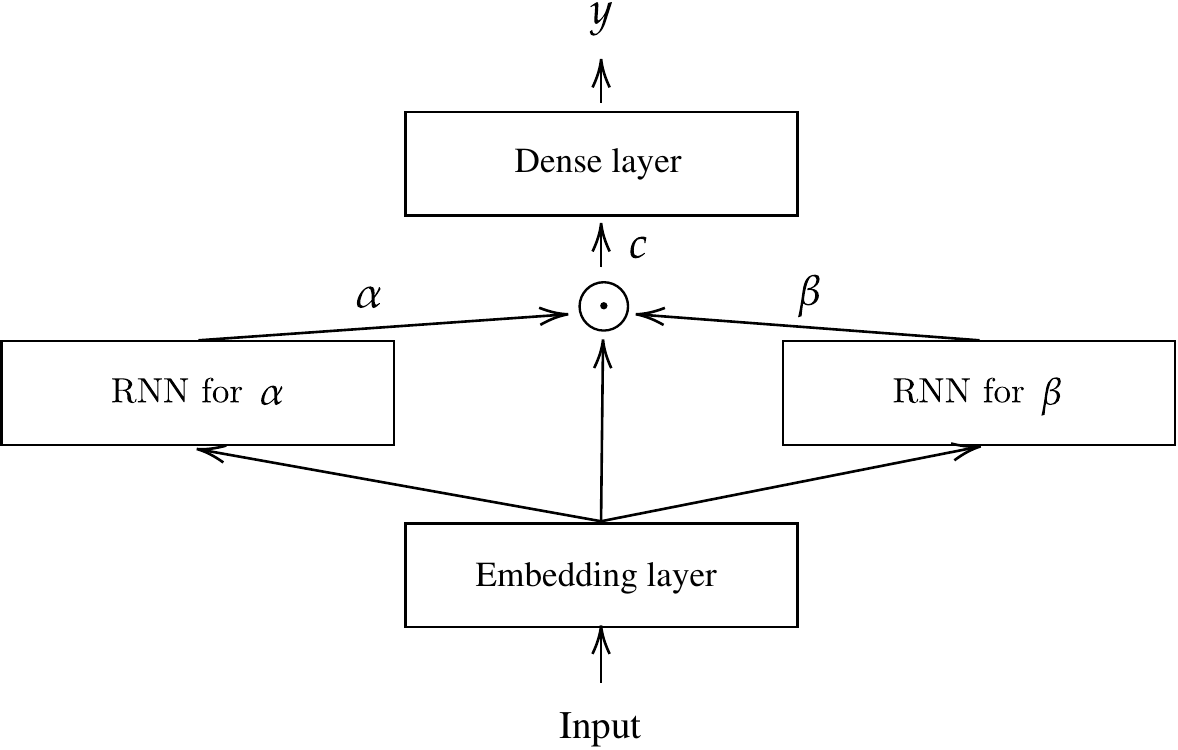}
\caption{Simplified architecture of RETAIN}
\label{fig:RETAIN}
\end{figure}

RETAIN is a neural network-based interpretable binary classification method developed by~\citet{choi2016retain}. 
The architecture restricts the form of prediction to be of Equation~\eqref{eqn:nlc} and defines contribution to the prediction of feature $f$ as $\gamma(x)_f \cdot x_f$.
The architecture uses two RNNs. The sigmoid function is applied over the sum of multiplication of the embedded input vectors and output of RNNs, to predict binary variable $y$, i.e., variable of interest. 
A simplified visualization of the RETAIN architecture is provided in Figure~\ref{fig:RETAIN}.

More formally, given the input data $(x_t)_{t \in [T]}$, where $x_t \in \mathbb{R}^p$ for all $t \in [T]$, $p$ being the number of features, the RETAIN architecture involves the following set of mathematical operations:
\begin{align}
    & v_i = W_{\text{emb}} x_i \\*[0.3em]
    & g_i, \hdots, g_1 = \text{RNN}_{\alpha}(v_i, \hdots, v_1) \\*[0.3em]
    & e_j = w_{\alpha}^\top g_j + b_{\alpha} & \text{for } j \in [i] \\*[0.3em]
    & \alpha_1, \hdots, \alpha_{T} = \text{Softmax}(e_1, \hdots, e_i) \\*[0.3em]
    & h_i, \hdots, h_1 = \text{RNN}_{\beta}(v_i, \hdots, v_1) \\*[0.3em]
    & \beta_j = \text{tanh}(w_{\beta}^\top h_j + b_{\beta}) & \text{for } j \in [i] \\*[0.3em]
    & c_t = \sum_{j \in [i]} \alpha_j \beta_j \odot v_j \\*[0.3em]
    & \mathbb{P}(y_i|(x_t)_{t \in [i]}) = \text{Softmax}(Wc_i + b) \label{eqn:RETAIN_out}
\end{align}
where $\alpha$ computes time step-wise attention scores and $\odot$ corresponds to the Hadamard product operator.

Let $r$ be the embedding dimension of the features. 
To define a feature level contribution for the output, \citet{choi2016retain} show that Equation~\eqref{eqn:RETAIN_out} can be expanded as follows:
\begin{align}
\begin{split}
  \mathbb{P}(y_i|(x_t)_{t \in [i]}) = \text{Softmax} \left( \sum_{j \in [i]} \sum_{k \in [r]} x_{j,k} \alpha_j W \left( \beta_j \odot W_{\text{emb}}[:, k] \right) + b \right)
\end{split}
\label{eqn:RETAIN_out_exp}
\end{align}
Then, based on Equation~\eqref{eqn:RETAIN_out_exp}, the contribution of feature $k$ at time step $j$ for the prediction of $y_i$ can be obtained as
\begin{align}
    \omega (y_i; x_{j,k}) = \alpha_j W (\beta_j \odot W_{\text{emb}})[:, k] x_{j,j}
\end{align}
Note that the definition of quantity $\omega (y_i; x_{j,k})$ is analogous to $\gamma^{\Theta}(x)$ component of the Equation~\eqref{eqn:nlc}, which can be used to explain why RETAIN is an interpretable model.

\subsection{Probabilistic parametric time series forecasting}
\label{subsec:probfor}
In general, time series forecasting methodologies can be classified as point forecasting and probabilistic forecasting. 
In point forecasting, the objective is to directly estimate the value for the subsequent steps of the target time series. 
On the other hand, in probabilistic forecasting, the probability of predicting a particular value is quantified for the target time series. 
More specifically, let $t$ be the time step and $I \in \mathbb{N}$ be the number of time series. Let $(x_{ij}, z_{i,j-1})_{t \in [I], j \in \{t-h, \cdots, t-1\}}$ such that $(x_{ij}, z_{i,j-1}) \in \mathbb{R}^{|\mathcal{F}|+1}$ is the collection of time series dataset, where $h$ is the history size.
We call $(x_{ij})_{ij}$ as covariates, and $(z_{ij})_{ij}$ as target series. 
In addition, assume that for all $i \in [I]$, $(x_{ij})_{t \in [t, T] \cap \mathbb{N}}$ are given, where $t+F > t$ with $F \in \mathbb{N}$. 
Then, the goal of probabilistic forecasting is to estimate the quantity given by
\begin{align}
    \mathbb{P}((z_{i,j})_{i \in [I], t \in [t, \hdots, t+F] \cap \mathbb{N}} | (z_{i,j})_{i \in [I], j \in \{t-h, \cdots, t-1\}}, (x_{ij})_{i \in [I], j \in \{t-h, \cdots, t + F\}})
    \label{eqn:probfor}
\end{align}
where $F$ is the forecasting time steps.

There are two main approaches to estimate the quantity shown in Equation~\eqref{eqn:probfor}:
\begin{itemize}
    \item \textit{Parametric} methods involve estimating $(z_{i,j})_{i \in [I], j \in \{t + 1, t+F\} \cap \mathbb{N}}$ by assuming $(z_{i,j})_{i \in [I], j \in \{t-h, \cdots, t-1\}}$ is a stochastic process from some joint density $\pi(\varsigma)$ with known $\pi$ and unknown $\varsigma$. As an example of parametric time series forecasting methodologies, we refer the readers to the work of \citet{salinas2020deepar}.
    
    \item \textit{Non-parametric} methods involve directly estimating Equation~\eqref{eqn:probfor} without assuming any knowledge about $\pi$. 
    For further details of such a non-parametric methodology, we refer readers to the works of \citet{koenker1982robust}, and \citet{lim2021temporal}. 
\end{itemize}

In this paper, we only focus on the parametric case. 
The primary benefit of proposing interpretable parametric time series forecasting methods is that it allows interpreting parameters of the stochastic process. 
For instance, by interpreting the parameters of a Gaussian process, we can achieve interpretation for both the mean value and standard deviation of the probabilistic forecasts. 

\section{\model Architecture}\label{sec:POLAR}
We next describe \model by detailing the model specifications and the architecture, and we discuss how the interpretability is achieved by this model.

\subsection{Model description}
\model is designed to perform joint density parameter estimation via performing time step-wise parameter forecasts. More precisely, we restrict the stochastic processes to be Gaussian, as it is a probability distribution which directly use the mean and standard deviation as its input. However, we note that \model architecture can be appropriately modified to handle other types of stochastic processes. 
For instance, following the steps proposed by \citet{salinas2020deepar}, such modifications can be easily achieved. 
For the sake of better readability, we eliminate the time series indices $i$ in below mathematical equations.



Firstly, let $M$ be number of continuous features, and $N$ be number of categorical features. 
We note that the input data can be decomposed as a vector containing both target series value, continuous features and discrete features. 
More precisely, input data can be represented as shown in Equation~\eqref{eqn:x_decom}, that is, for all $j \in \{t-h, \cdots, t-1\}$, where $h$ is history size, time series with covariate can be written as
\begin{align}
    [z_j, (x^{\text{Cont}}_{jf})_{f=1}^M, (x^{\text{Cat}}_{jf})_{f=1}^N]
    \label{eqn:x_decom}
\end{align}
where $z_j$ is target series value at time step $j$, $(x^{\text{Cont}}_{jf})_{f=1}^M$ denotes vector of continuous components of vector $x$, while $(x^{\text{Cat}}_{jf})_{f=1}^N$ denotes the categorical components. For the sake of notational convenience, we make the following assumptions:
\begin{itemize}
    \item Without loss of generality, we assume such an index ordering as default.
    
    \item We let $\mathcal{F}_{\text{Cat}}$ be set of categorical feature indices of the vector shown in Equation~\eqref{eqn:x_decom}. Analogously, we define $\mathcal{F}_{\text{Cont}}$ for continuous features. 
    
    \item We assume $x^{\text{Cat}}_{jf} \in \mathbb{N}$, i.e., categories are represented in terms of natural numbers. We assume that the values of $(x^{\text{Cat}}_{itf})_{it}$ lie in a subset of ordered natural numbers without any gap starting from 0, so the values of categorical features can be one-hot encoded.
\end{itemize}

\begin{figure}[!ht]
\centering
\includegraphics[width=0.75\columnwidth]{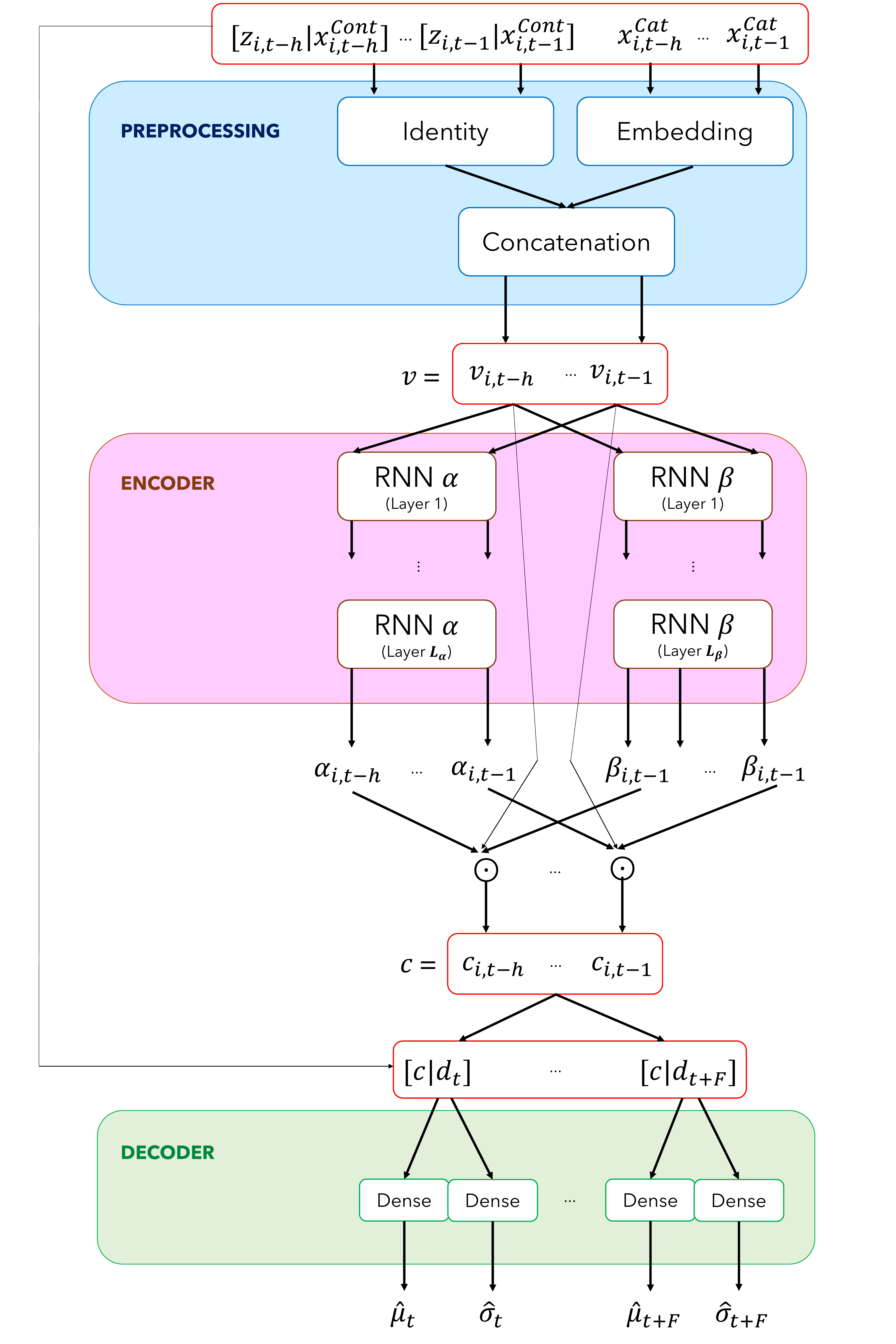}
\caption{\model model architecture}
\label{fig:POLAR}
\end{figure}

\noindent The \model is defined as concatenation of three important parts: feature processing layer, encoding layer and decoding layer. In Figure~\ref{fig:POLAR}, we present a simplified overview of the entire neural network architecture for \model. 

We start the model description by depicting feature processing layer. For each time step $j \in \{t-h, t-1\}$,
\begin{align}
    & e^{\text{Cat}}_{f} = \text{Embedding} (x^{\text{Cat}}_{jf}) & \forall f \in \mathcal{F}_{\text{Cat}} \\*[0.3em] 
    & v_{j} = (z_j, x^{\text{Cont}}_{j,1} \cdots x^{\text{Cont}}_{j,M}, e_{j,1} \cdots e_{j,N})
\end{align}
where for all $f \in \mathcal{F}_{\text{Cat}}$, $e_{jf}^{\text{Cat}} \in \mathbb{R}^{\xi_f}$ with $\xi_f$ representing the embedding size of categorical feature $f \in \mathcal{F}_{\text{Cat}}$. 
Using notations above, we define the input for the encoding layer as
\begin{align}
    v = [v_{t-h}, \cdots, v_{t-1}]  
\end{align}
Then we can subsequently define the encoding layer as follows:
\begin{align}
    & g_{t-h}, \hdots, g_{t-1} = \text{RNN}_{\alpha}(v_{t_h}, \hdots, v_{t_1}) & \\*[0.3em]
    & q_{j} = w_{\alpha}^{\top}g_{j} + b_{\alpha} &  \forall j \in \{ t-h, \cdots, t-1\} \\*[0.3em] 
    & \alpha_{t-h}, \hdots, \alpha_{t-1} = \text{Softmax}(q_{t-h}, \hdots, q_{t-1}) &  \\*[0.3em] 
    & h_{t-h}, \hdots, h_{t-1} = \text{RNN}_{\beta}(v_{t-h}, \hdots, v_{t-1}) &  \\*[0.3em]
    & \beta_{j} = \text{tanh}(w_{\beta}^{\top}h_{j} + b_{\beta}) &  \forall j \in \{ t-h, \cdots, t-1\} \\*[0.3em] 
    & c_{j} = \alpha_{j} \beta_{j} \odot v_{j} &  \forall j \in \{ t-h, \cdots, t-1\}
\end{align}
where both $\text{RNN}_{\alpha}(\cdot)$ and $\text{RNN}_{\beta}$ can be either uni- or bi-directional.\\

The output of the encoding layer is
\begin{align}
    c = (c_{t-h}, \cdots, c_{t-1})
\end{align}
which can be used to define the input for the decoding layer as
\begin{align}
    & d_{j} = [z_{j-1}, x^{\text{Cont}}_{j,1} \cdots x^{\text{Cont}}_{j,M}, e_{j,1} \cdots e_{j,N}]&  \forall j \in \{ t, \cdots, t+F\} \\*[0.3em]
    & d = (d_t, \cdots, d_{t+F}) &
\end{align}
Using the sequence of vectors $(c_{j})_{j \in \{t-h, \cdots, t-1\}}$, we can finally perform mean and standard deviation prediction in the decoding layer by using the following formula:
\begin{align}
    & \mu_{j} = W_{\mu,j}^{\top} \cdot [c|d_{j}]&  \forall j \in \{ t, \cdots, t+F\}\\*[0.3em]
    & \sigma_{j} = \text{Softplus}(W_{\sigma,j}^{\top} \cdot [c|d_{j}]) &  \forall j \in \{ t, \cdots, t+F\}
\end{align}
where $W_{\mu,j}, W_{\sigma,j} \in \mathbb{R}^{1 \times N \times \sum_{f \in \mathcal{F}} d_f}$. 
We note that both $\mu_{it}$ and $\sigma_{it}$ are later used to define the parameters for output conditional univariate Gaussian distribution. 

Given the neural network output, the neural network weights are optimized to solve the likelihood maximization problem given by
\begin{align}
    \max_{\Theta}  \prod_{i \in I} \prod_{j =t}^{t+F} \ell_{\mathsf{UVN}}(z_{i, j} | \mu_{ij}(\Theta), \sigma_{ij}(\Theta))
\end{align}
where, instead, we solve the log-likelihood maximization problem described by Equation~\eqref{eqn:maxlik}, that is,
\begin{align}
     \max_{\Theta}  \sum_{i \in I} \sum_{j =t}^{t+F} \log (\ell_{\mathsf{UVN}}(z_{i, j} | \mu_{ij}(\Theta), \sigma_{ij}(\Theta)))
     \label{eqn:maxlik}
\end{align}
We recover the eliminated index $i$ to describe the time series indices. 
We use $\ell_{\mathsf{UVN}}(\cdot|\cdot)$ to denote univariate Gaussian likelihood, where $\Theta$ is the vector of all network parameters optimized according to the network inputs.

\begin{figure}[!ht]
\centering
\includegraphics[width=0.75\columnwidth]{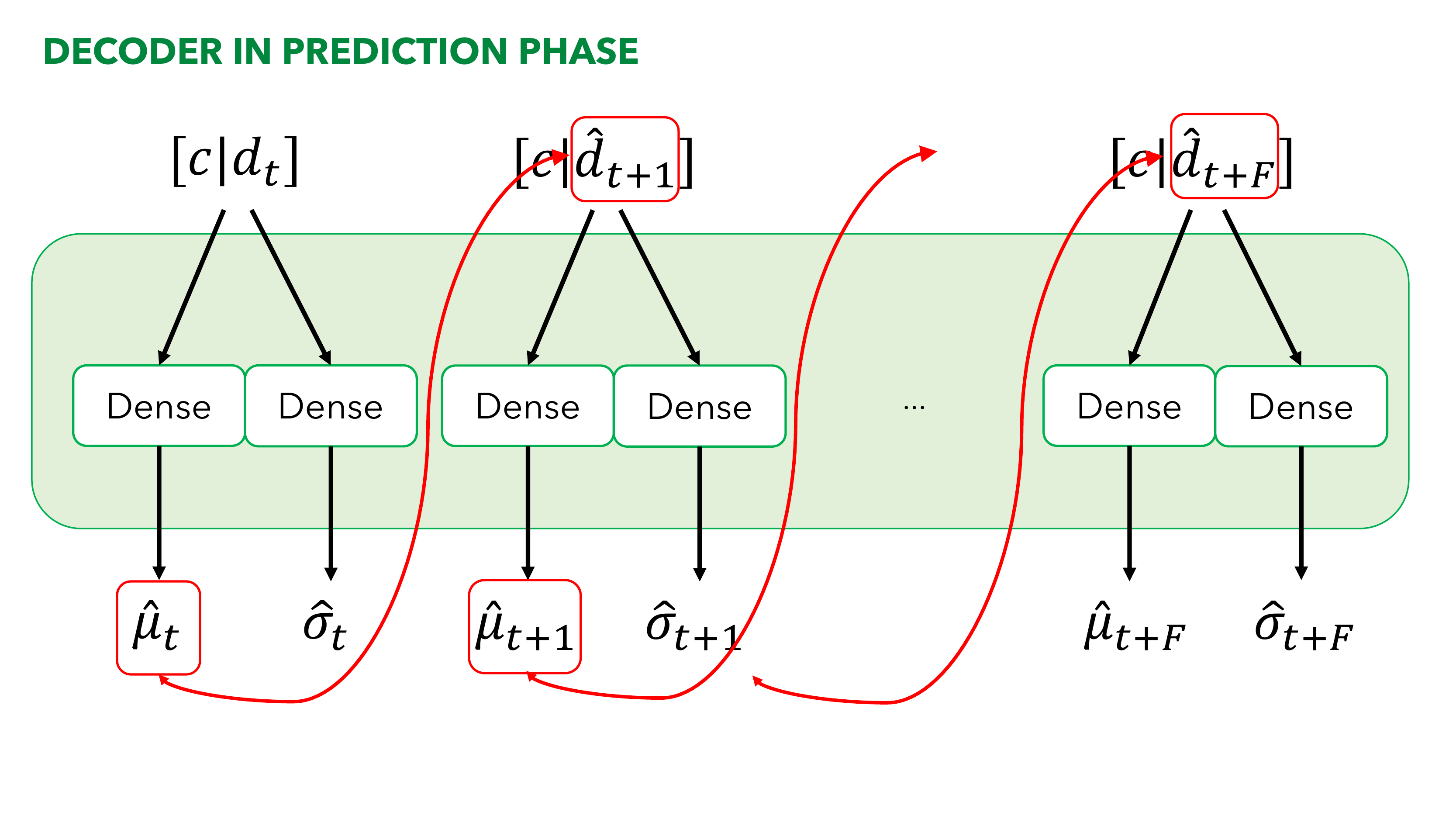}
\caption{Decoder of the \model architecture during forecasting phase} 
\label{fig:POLAR_PRED}
\end{figure}

During the training of the \model, the input for decoding layer is defined using true target series value $z_j$ for each $j \in \{ t-h, \cdots, t-1\}$. However, during the prediction phase, the future information $z_j$ for $j \in \{ t, \cdots, t+F\}$ are not accessible. 
To overcome this issue, we perform autoregressive prediction, and we replace the decoding layer input vector $d_j$ by $\hat{d}_j$. Specifically,
\begin{align}
    & d_{t} = [z_{t-1}, x^{\text{Cont}}_{t,1} \cdots x^{\text{Cont}}_{t,M}, e_{t,1} \cdots e_{t,N}] \\
    & \hat{d}_{j} = [\hat{z}_{j-1}, x^{\text{Cont}}_{j,1} \cdots x^{\text{Cont}}_{j,M}, e_{j,1} \cdots e_{j,N}]&  \forall j \in \{ t+1, \cdots, t+F\} \\
    & \hat{d} = (d_t, \hat{d}_{t+1}, \cdots, \hat{d}_{t+F}) &
\end{align}
where $\hat{z}_{j-1}$ is a sample drawn from $\pi(\hat{\mu}_{j-1}, \hat{\sigma}_{j-1})$, which denotes the predicted target series value for all $j \in \{ t+1, \cdots, t+F\}$.
During the forecasting phase, \model performs sequence sampling as shown in Figure~\ref{fig:POLAR_PRED}.
\subsection{Model interpretation}\label{sec:csc}


We can assess how much the input features contribute to the model output when the $\model$ makes a prediction. 
For the sake of brevity, let $\phi = 1+|\mathcal{F}^{\text{Cont}}| + \sum_{r=0}^{f} \xi_r$ be the length of the one time step context vector.
We can define the contribution of feature $f$ at time step $s$ for $\mu_j$ prediction as follows:
\begin{footnotesize}
\begin{align}
    \nonumber \omega_{\mu}(f, s, j) &= \alpha_{s} W_{\mu,j}[(s-(t-h)) \cdot \phi] \cdot (\beta_{s}[0] \cdot z_{s}),\\ 
    & \label{eqn:c_mu1} \-\hspace{7.09cm} f \notin \mathcal{F}^{\text{Cont}} \cup \mathcal{F}^{\text{Cat}} \\*[0.6em]
    \nonumber \omega_{\mu}(f, s, j) &= \alpha_{s} W_{\mu,j}[(s-(t-h)) \cdot \phi + f] \cdot (\beta_{s}[f] \cdot x_{sf}),\\
    & \label{eqn:c_mu2} \-\hspace{7.59cm} f \in \mathcal{F}^{\text{Cont}}\\*[0.6em]
    \nonumber \omega_{\mu}(f, s, j) &= \alpha_{s} W_{\mu,j} \bigg[(s-(t-h)) \cdot \phi + (\phi - \xi_f):(s-(t-h)) \cdot \phi + 
    \phi \bigg] \\*[0.3em]
    \nonumber & \-\hspace{-1.2cm} \cdot \Big(\beta_{s}\bigg[ (s-(t-h)) \cdot \phi + (\phi - \xi_f): (s-(t-h)) \cdot \phi + \phi
    \bigg] \odot W^{\text{Cat,Emb}}_f[:,x_{sf}]\Big),\\*[0.3em]
    & \label{eqn:c_mu3} \-\hspace{7.59cm} f \in \mathcal{F}^{\text{Cat}}
\end{align}
\end{footnotesize}

Similarly, the contribution for $\sigma_j$ prediction is defined as 
\begin{footnotesize}
\begin{align}
    \nonumber \omega_{\sigma}(f, s, j) &= \alpha_{s} W_{\sigma,j}[(s-(t-h)) \cdot \phi] \cdot (\beta_{s}[0] \cdot z_{s}),\\ 
    & \label{eqn:c_sigma1} \-\hspace{7.09cm} f \notin \mathcal{F}^{\text{Cont}} \cup \mathcal{F}^{\text{Cat}} \\*[0.6em]
    \nonumber \omega_{\sigma}(f, s, j) &= \alpha_{s} W_{\sigma,j}[(s-(t-h)) \cdot \phi + f] \cdot (\beta_{s}[f] \cdot x_{sf}),\\
    & \label{eqn:c_sigma2} \-\hspace{7.59cm} f \in \mathcal{F}^{\text{Cont}}\\*[0.6em]
    \nonumber \omega_{\sigma}(f, s, j) &= \alpha_{s} W_{\sigma,j} \bigg[(s-(t-h)) \cdot \phi + (\phi - \xi_f):(s-(t-h)) \cdot \phi + 
    \phi \bigg] \\*[0.3em]
    \nonumber & \-\hspace{-1.2cm} \cdot \Big(\beta_{s}\bigg[ (s-(t-h)) \cdot \phi + (\phi - \xi_f): (s-(t-h)) \cdot \phi + \phi
    \bigg] \odot W^{\text{Cat,Emb}}_f[:,x_{sf}]\Big),\\*[0.3em]
    & \label{eqn:c_sigma3} \-\hspace{7.59cm} f \in \mathcal{F}^{\text{Cat}}
\end{align}
\end{footnotesize}

We note that the provided formula is valid, as long as all continuous features come before categorical features, and target series come before all continuous features. Without such an assumption, the formula for the contribution score must be appropriately modified.
When $\omega(f, s, j) > 0$ the contribution of feature $f$ at timestep $j$ is positive. This means the provided component contributes to an increase in the value of the prediction. 
On the other hand, a negative value means that the contribution score is negative. 
Without the monotonically increasing property of the activation functions of the dense layers in the decoder, such an interpretation would not be possible. 
It is also important to note that the contributions of features on standard deviation are nonlinear, which is due to the existence of the Softplus operator.
By taking the norm of the contribution scores $|\omega(f, s, j)|$, we can obtain the importance of the features, instead of contribution scores.
These features can then be averaged over the prediction steps and samples of the dataset to obtain the average importance of the input features over the dataset, which enables the global-level contribution score computation for the features.

\section{Numerical Study}\label{sec:exp}
In this section, we present the results from detailed experiments to evaluate the forecasting performance and interpretability of the proposed model, \model. 
We first provide the experimental setup and detail the considered forecasting models, hyperparameters and datasets.
Then, we compare the forecasting performance of several models using well-known point and probabilistic forecasting evaluation metrics. 
Next, we focus on the results that demonstrate the interpretability aspects of \model. 
Specifically, we first show the explanations produced by the model, and then discuss these explanations as they relate to the underlying prediction task.

\subsection{Experimental setup}
We consider three diverse datasets with different characteristics in terms of size, and observed seasonality/trends among others.
We use a sliding window method for framing the datasets, and separate the last two forecasting horizons of each time series in the datasets as validation and test sets. 
We perform a detailed hyperparameter tuning for all models using Tree-structured Parzen Estimator (TPE) algorithm. 
The parameter ranges for the hyperparameter tuning are shown in Table~\ref{tab:hyperparameters_space}. 
For \model and DeepAR, we experiment with different number of RNN layers, RNN cell types, hidden units, dropout and learning rates. 
For the MLP model, we experiment with a similar range of hidden layers, learning and dropout rates. 
For hidden units, we use a wider range, as dense neural networks are usually trained with a larger number of hidden units to achieve similar performance compared to the RNN based networks.
For the GBR model, we test three important parameters which are the number of trees in the ensemble, the number of leaves in each tree (i.e., \textit{max depth}), and the minimum number of samples required to split an internal node (i.e., \textit{min samples split}).

\setlength{\tabcolsep}{4.5pt} 
\renewcommand{\arraystretch}{1.23} 
\begin{table}[!ht]
    \centering
    \caption{The hyperparameter tuning search space.}
    \label{tab:hyperparameters_space}
    \resizebox{0.637\textwidth}{!}{
        \begin{tabular}{|l|ll|}
        \hline
        \textbf{Model} &  & \textbf{Search space} \\
        \hline
        \multirow[t]{3}{*}{\model / DeepAR} & & \makecell[l]{ \textit{\# Hidden units}: [16, 128],\\ \textit{Dropout rate}: [0, 0.5],\\ \textit{Cell Type:} LSTM or GRU,\\ \textit{\# RNN layers}: [1, 8],\\ \textit{Learning Rate}: [1e-4, 1e-1] (log uniform)}\\*[0.6em]
        \hline
        \multirow[t]{3}{*}{MLP} & & \makecell[l]{ \textit{\# Hidden units}: [50, 500],\\ \textit{Dropout rate}: [0, 0.5],\\ \textit{\# Hidden layers}: [1, 8],\\ \textit{Learning Rate}: [1e-4, 1e-1] (log uniform)}\\
        \hline
        GBR & & \makecell[l]{\textit{\# of trees}: [10, 200],\\ \textit{Max depth}: [2, 5],\\ \textit{Min samples split}: [2, 15]}\\
        \hline
        \end{tabular}
    }
\end{table}


We run the TPE algorithm for 100 trials, optimizing over the normalized deviation metric on the validation set.
The neural network models are implemented using Pytorch, the GBR model is implemented using Scikit-learn, and hyperparameter tuning is performed using the Optuna library. 
All the experiments are run on a computing node with RTX2070 Super 8GB GPU, and 128GB of RAM, running on Debian Linux OS.

We consider two sets of performance metrics to evaluate both point and probabilistic forecasting performance.
For measuring the quality of point forecasts, we use Normalized Root Mean Squared Error (NRMSE) and Normalized Deviation (ND) similar to~\citet{salinas2020deepar}. 
NRMSE and ND can be obtained for ground truth values ($y$) and the prediction ($\hat{y}$) as follows:
\begin{align}
    \text{NRMSE}(y, \hat{y}) = \frac{\sqrt{\frac{1}{N} \sum_{i=1}^{N}(\hat{y}_i - y_i)^2}}{\frac{1}{N} \sum_{i=1}^N\vert y_i\vert}, \qquad
    \text{ND}(y, \hat{y}) = \frac{\sum_{i=1}^{N}\vert\hat{y}_i - y_i\vert}{ \sum_{i=1}^{N} \vert y_i \vert}
    \label{eqn:NRMSE_ND}
\end{align}

To evaluate the probabilistic forecasting ability of the models, we consider the $\rho$-risk, also known as quantile loss. The $\rho-$risk can be obtained as follows:
\begin{align}
    \rho_{\alpha}(y, \hat{y}^{\alpha}) &= \frac{\sum_{i=1}^{N} \max\{\alpha(y_i - \hat{y_i}^\alpha), (1-\alpha)(\hat{y_i}^\alpha - y_i)\}}{\sum_{i=1}^{N} \vert y_i \vert}
\end{align}


\noindent where $\hat{y}^\alpha$ represents the prediction at quantile level $\alpha$.
Note that having a lower value of $\rho$-risk is better. 
Finally, we apply Friedman test, a non-parametric test, to compare the forecasting models on multiple datasets~\citep{garcia2010advanced}. 
For this comparison, we fill each group with the average error results on the tested datasets.
Then, we apply the Friedman test to find out whether there is a statically significant difference between the mean of the groups. 

We consider three datasets in our experiments, namely, Electricity, Rossmann and Walmart datasets, which we briefly describe below.
\begin{itemize}
    \item \textit{Electricity}: The Electricity dataset contains the hourly electricity consumption records of 370 households. 
    The dataset has been extensively used in previous studies for performance benchmarking purposes \cite{lim2021temporal, salinas2020deepar}. 
    In our experiments, besides the electricity consumption time series, we use additional covariates such as \textit{hour of the day}, \textit{day of the week}, \textit{week of the month}, and \textit{month}.
    Input window size and forecasting horizon are taken as 168 and 12, respectively. 
    
    \item \textit{Rossmann}: The Rossmann sales dataset was published as a part of a Kaggle competition in 2015 and it contains a rich set of features and extensive daily sales histories. 
    The dataset consists of daily sales records of multiple Rossmann stores, with various covariates. 
    Among the available features, we use \textit{sales value}, \textit{store index}, \textit{store open indicator}, \textit{promotion indicator}, \textit{state holiday indicator}, \textit{school holiday indicator}, \textit{weekday}, \textit{month}, and \textit{week of the month}.
    Input window size and forecasting horizon are taken as 30 and 12, respectively.

    \item \textit{Walmart}: The Walmart store sales forecasting dataset contains weekly store sales of 77 departments in 45 stores, and it was made available as part of a Kaggle competition in 2014.
    The dataset includes multiple features such as \textit{temperature}, \textit{information on markdowns}, and \textit{various economic indicators} (e.g., unemployment rates, fuel prices and CPI). 
    Input window size and forecasting horizon are taken as 30 and 6, respectively.
\end{itemize}

\subsection{Results on model performances}
We provide summary statistics on the model performances in Table~\ref{tab:results_table}. 
These results are obtained after performing rigorous fine-tuning for each dataset-model pair on the validation set.
The Friedman test over the aggregate results (i.e., results for all three datasets combined) returns $p$-values of 0.12, 0.08, 0.12, and 0.12 for NRMSE, ND, $\rho_{0.75}$, and $\rho_{0.90}$, respectively, which indicates that there is no statistically significant difference ($p$-value $>$ 0.05) between \model and DeepAR models. 
Similarly, we see that \model and DeepAR achieve similar forecasting performance in terms of average performance values. 
\model performs marginally better for the Rossmann and Electricity datasets, whereas DeepAR performs marginally better for the Walmart dataset.

\renewcommand{\arraystretch}{1.2} 
\begin{table}[hbt!]
\centering
\caption{ 
Summary statistics of model performances. Mean and standard deviation across 10 randomly seeded runs are reported over the test sets.}
\label{tab:results_table}
\resizebox{0.825\textwidth}{!}{
\begin{tabular}{lrcccccccc}
\midrule
\multirow{2}{*}{\textbf{Dataset}} & \multirow{2}{*}{\textbf{Model}} & \multicolumn{2}{c}{\textbf{NRMSE}} & \multicolumn{2}{c}{\textbf{ND}} & \multicolumn{2}{c}{$\rho_{0.75}$} & \multicolumn{2}{c}{$\rho_{0.90}$}\\
\cline{3-10}
 & & \textbf{mean} & \textbf{std} & \textbf{mean} & \textbf{std} & \textbf{mean} & \textbf{std} & \textbf{mean} & \textbf{std} \\ 
\midrule
\multirow[t]{4}{*}{\textbf{Electricity}} & DeepAR &  0.255 & 0.017 &    0.077 & 0.005 &         0.133 & 0.026 &        0.250 & 0.041 \\
        & GBR &       0.223 & 0.002 &    0.066 & 0.000 &         0.116 & 0.000 &        0.229 & 0.002 \\
        & MLP &       0.359 & 0.043 &    0.120 & 0.010 &         0.238 & 0.019 &        0.451 & 0.035 \\
        & \model &       0.221 & 0.020 &    0.070 & 0.004 &         0.130 & 0.017 &        0.249 & 0.027 \\

\midrule
\multirow[t]{4}{*}{\textbf{Rossmann}} & DeepAR &       0.177 & 0.027 &    0.119 & 0.020 &         0.241 & 0.070 &        0.463 & 0.121 \\
        & GBR &       0.148 & 0.000 &    0.102 & 0.000 &         0.195 & 0.000 &        0.483 & 0.001 \\
        & MLP &       0.179 & 0.005 &    0.122 & 0.003 &         0.281 & 0.013 &        0.596 & 0.032 \\
        & \model &       0.153 & 0.005 &    0.105 & 0.004 &         0.214 & 0.027 &        0.407 & 0.056 \\

\midrule
\multirow[t]{4}{*}{\textbf{Walmart}} & DeepAR &       0.152 & 0.013 &    0.076 & 0.007 & 0.136 & 0.026 & 0.248 & 0.043 \\
        & GBR &       0.198 & 0.005 &    0.093 & 0.001 &         0.164 & 0.007 &        0.363 & 0.025 \\
        & MLP &       0.228 & 0.023 &    0.120 & 0.013 &         0.250 & 0.037 &        0.465 & 0.062 \\
        & \model &       0.186 & 0.010 &    0.095 & 0.006 &         0.151 & 0.032 &        0.307 & 0.061 \\
\bottomrule
\end{tabular}
}
\end{table}


Baseline models, MLP and GBR, are significantly less complex compared to DeepAR and \model.
MLP leads to the poorest forecasting performance among the tested models for all the datasets as shown by various forecasting performance values. 
GBR outperforms \model and DeepAR on the Rossmann dataset, where the dataset shows a significant amount of seasonality. 
For the Electricity dataset, GBR ranks second in terms of the NRMSE metric, and first in terms of the ND metric, with a marginal difference in both cases. 
For the Walmart dataset, GBR ranks third after DeepAR and \model in terms of the NRMSE metric, and second in terms of the ND metric. 
The high performance of the GBR can be attributed to the way we train this model.
Specifically, for the GBR model, we train a separate model for each time series on the dataset, whereas, for the other models, we train a single model for all the time series in the dataset. 
This training methodology for GBR is computationally more expensive. 
However, it can result in a significantly improved forecasting performance when the availability of similar time series does not benefit the prediction.
Our preliminary analysis indicates that training a single GBR model for each dataset (i.e., similar to the other three models) leads to a significant deterioration in forecasting performance across the datasets, hence we adopt the above-explained approach.
We note that above results for these datasets are largely inline with the forecasting performance values reported in previous studies (e.g., see \citep{ilic2021explainable, ozyegen2020evaluation, salinas2020deepar}).

\begin{figure}[!ht]
    \centering
        \subfloat[Rossmann: DeepAR ]{\includegraphics[width=0.5\textwidth]{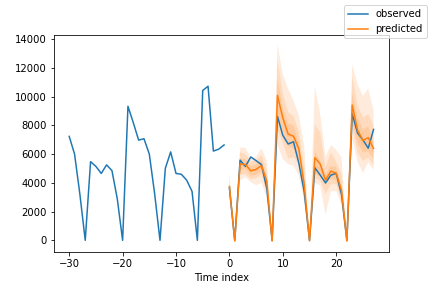}
        \label{fig:ross_deepar_pred_vis}
        }
        \subfloat[Rossmann: \model ]{\includegraphics[width=0.5\textwidth]{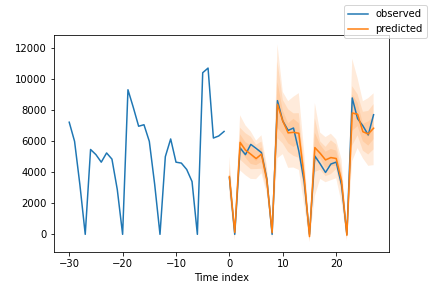}
        \label{fig:ross_pidits_pred_vis}
        }\\
        \subfloat[Electricity: DeepAR ]{\includegraphics[width=0.5\textwidth]{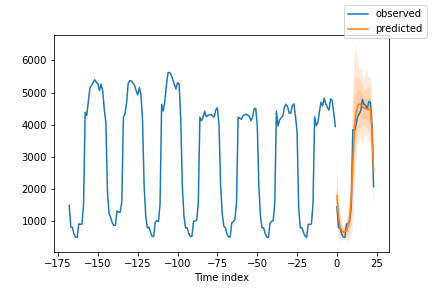}
        \label{fig:elect_deepar_pred_vis}
        }
        \subfloat[Electricity: \model ]{\includegraphics[width=0.5\textwidth]{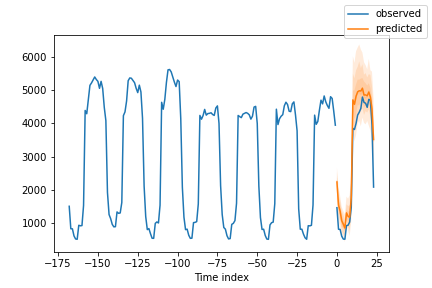}
        \label{fig:elect_pidits_pred_vis}
        }\\
        \subfloat[Walmart: DeepAR ]{\includegraphics[width=0.5\textwidth]{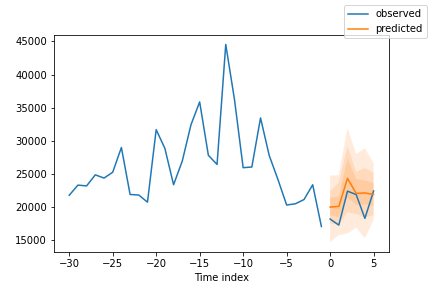}
        \label{fig:walmart_deepar_pred_vis}
        }
        \subfloat[Walmart: \model ]{\includegraphics[width=0.5\textwidth]{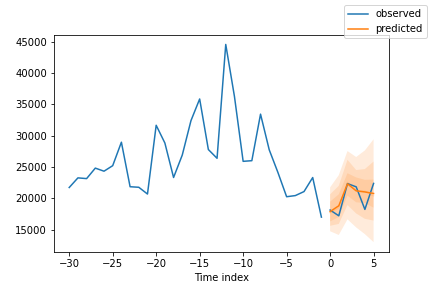}
        \label{fig:walmart_pidits_pred_vis}
        }\\
        
    \caption{Visualization of DeepAR and \model predictions on random time series samples. Each figure displays forecasts for one prediction window. Point forecasts are shown in the orange line, 75\%, 90\% and 98\% quantile forecasts are shown as orange regions.
    Each model generates predictions that can capture the trends for the provided sample.}
    \label{fig:visualize_pred}

\end{figure}

Figure~\ref{fig:visualize_pred} shows the visualization of forecasts by DeepAR and \model for three datasets. 
We use blue lines to represent the ground truth, and orange lines to denote the predicted mean values of models. 
We depict 75\%, 90\% and 98\% prediction intervals as shaded areas. 
The forecasting horizons are 28 days, 24 hours, and 6 weeks for Rossmann, Electricity, and Walmart datasets, respectively. 
Overall, we observe that all the models can generate predictions that capture the general trends in the datasets and probabilistic forecasts are highly similar for these two models.





\subsection{Results on model interpretability}
We next provide results on the interpretability of \model.
We focus on the Rossmann dataset as the representative case, as it has a large, interesting set of features that correlate well with the target variable (e.g., Promo). 
For the Electricity dataset, we see that the target value (i.e., electricity load) has the most impact on the predictions, and certain time covariates such as ``week-of-month'' help reduce the variance of the predictions. 
For the Walmart dataset, we again see that the target value (i.e., weekly sales) has the most impact on the predictions. 
The interpretability results for the Walmart and Electricity datasets are provided in the Appendix (see~\ref{appendix:additional_results_interpretability}).

We discuss the interpretability of \model by analyzing the explanations obtained from the model weights. 
Following the similar work~\cite{guo2019exploring, norgeot2018time}, we visualize the explanations as feature contribution heatmaps. 
The contribution score formulas described in the methodology are applied to obtain the contribution of each input feature to the prediction. 
Note that we collect a separate contribution score for each sample, input feature, and forecasting horizon. 
This allows us to achieve \textit{local interpretability}, which helps understanding how the features contribute to a single prediction. 
It is also possible to aggregate these scores over forecasting horizons and samples to achieve \textit{global interpretability}, which helps understand the features that are important for the model. 

\begin{figure}[hbt!]
    \centering
        \subfloat[$\mu$ contributions ]{\includegraphics[width=0.5\textwidth]{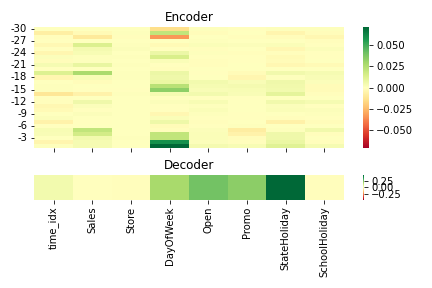}
        \label{fig:rossmann_sample0_mu}
        }
        \subfloat[$\sigma$ contributions]{\includegraphics[width=0.5\textwidth]{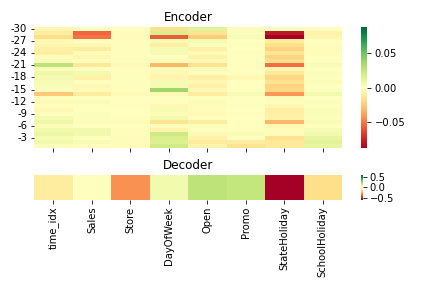}
        \label{fig:rossmann_sample0_sigma}
        }\\
    \caption{Contribution scores obtained from \model for the first sample and 9th target horizon of the Rossmann test set. Figure (a) shows the contributions to the $\mu$ and (b) shows contributions to the $\sigma$.}
    \label{fig:rossmann_sample_cont}

\end{figure}

Figure~\ref{fig:rossmann_sample_cont} shows the contributions scores obtained from the model for a single sample. 
We select the 9th forecasting horizon of the first sample from the Rossmann test set to generate these visualizations. 
This sample and the forecasting horizon provide a clear overview of how the model behaves under certain events. 
The forecasted date for this sample is a Monday, the store is Open, there is a promotion event, and there is no state or school holiday.

Figure~\ref{fig:rossmann_sample0_mu} shows that `DayOfWeek', `Open', `Promo', and `StateHoliday' have the biggest positive contribution to the prediction. 
This is intuitive since there are no sales on the weekends, and Mondays should expect a higher sale. 
The store being open, the promotion events and the state holidays are all positive contributors to the prediction. 
Looking at the encoder contributions, we observe that the encoder features have negligible impact on the prediction. 
However, we also note that recent and certain seasonal values of the `DayOfWeek' feature have higher impacts on predictions compared to the other encoder features.

Figure~\ref{fig:rossmann_sample0_sigma} shows the contribution of the features to the predicted variance. 
Overall, we observe that the `StateHoliday`, `Store', and `SchoolHoliday' decoder features have significant negative contributions, and `Open' and `Promo' have positive contributions to the prediction. 
The negative values for the store and holiday features show that they reduce the variance of the predictions. 
Interestingly, `StateHoliday' has the largest negative contribution to the variance. 
Since there is no date corresponding to a holiday in this example, the explanation may indicate that state holidays can result in a higher uncertainty (variance), which requires further investigation.
On the other hand, we observe a positive contribution for `Open', and `Promo' when the store is open and there is a promotion. 
This is expected as there is higher uncertainty in the predicted sales when the store is open as opposed to the days when the store is closed and there are no sales. 
The explanation also suggests that the existence of promotion has a similar impact on the prediction, and it causes a higher variance.

\begin{figure}[hbt!]
    \centering
        \subfloat[$\mu$ average importance ]{\includegraphics[width=0.5\textwidth]{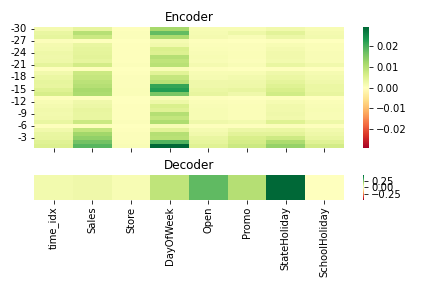}
        \label{fig:rossmann_avg_mu}
        }
        \subfloat[$\sigma$ average importance]{\includegraphics[width=0.5\textwidth]{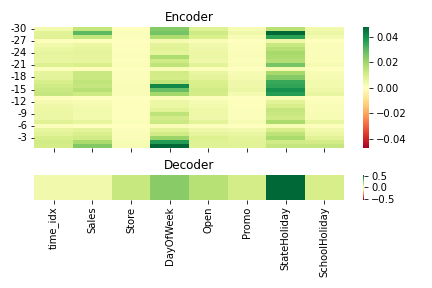}
        \label{fig:rossmann_avg_sigma}
        }\\
    \caption{Contribution scores obtained from \model for all samples, and averaged over samples to obtain global importance scores. Figure (a) shows the average importance to the $\mu$ and (b) shows average importance to the $\sigma$ in Rossmann dataset.}
    \label{fig:rossmann_avg_imp}
\end{figure}

In certain cases, it might be useful to calculate the overall importance of the features for the model, instead of their importance for a particular prediction. 
These feature importance values can then be used for different purposes including feature selection.
We apply the following steps to the contribution scores to find the overall importance. 
First, we obtain contribution scores for each sample in the dataset. 
The resulting array has the shape: $\#forecasting\_horizon \times \#samples \times \#timesteps \times \#features$. 
We then take the absolute values of this array, as we are interested in the overall importance of the features, not their contributions. 
Finally, we average the array over the forecasting horizons, and then over the samples to obtain the \textit{importance scores} of the input features. 
We show the visualizations of these arrays for $\mu$ and $\sigma$ in Figure~\ref{fig:rossmann_avg_mu} and Figure~\ref{fig:rossmann_avg_sigma}, respectively. 
These figures show the overall importance of the features for the model. 
Thus, all the values displayed are positive numbers, and they are displayed in green color. 
Looking at the importance scores for $\mu$ in Figure~\ref{fig:rossmann_avg_mu}, we observe that the same four decoder features from the sample contributions (`DayOfWeek', `Open', `Promo', and `StateHoliday') are found to be important for the model. 
Additionally, we note that various encoder timesteps of the `Sales', and `DayOfWeek' features are also important, suggesting that the model attends to different timesteps of these features for different samples of the dataset. 
Overall importance scores for $\sigma$ shown in Figure~\ref{fig:rossmann_avg_sigma} follow a similar pattern compared to the sample contribution scores. 
The same decoder features, i.e., `Stateholiday', `Store', `Open', `Promo', and `SchoolHoliday' are all important for the model. 
Additionally, the `DayOfWeek' feature of the decoder is also important for predicting the target value. On the decoder side, certain past values of `Sales', `DayOfWeek' and `StateHoliday' are all found to be important to the variance prediction.

\section{Conclusion and Discussions}\label{sec:conclusions}
In this paper, we present a novel deep probabilistic intrinsically interpretable time series forecasting method, \model, which is designed to predict both mean and standard deviation of the underlying Gaussian processes. 
We show how ordinal and categorical covariates can be appropriately incorporated into \model. 
Then, we describe procedures to compute the contribution of both ordinal and discrete features using \model network parameters. 
We also compare \model against various baselines and show that our method has a competitive forecasting performance to DeepAR, a state-of-the-art probabilistic time series forecasting method. 
We also analyze the contribution scores generated for mean and standard deviation. 
We show how contribution scores can be used to analyze model predictions for a single sample and to find the overall importance of the features for the model. 

An area that could be further explored is the idea of using the contribution scores for feature selection. 
Because the contribution scores are calculated directly from the model weights, they may achieve a higher performance in feature selection compared to the alternative methods.
Secondly, the proposed architecture can be tested with other datasets. 
While \model achieves similar performance to DeepAR for the tested datasets, the architecture might require further tuning for other datasets and problems. 
Finally, the role of regularization can be analyzed in finding the most useful explanations. 
Methods such as L1 regularization and Dropout can be used to tweak the contributions assigned to the correlated features. 
Depending on the use case, certain regularization hyperparameters can be selected to obtain the most useful explanation.


\section*{Acknowledgment}
This research is in part supported by LG Science Park.

\section*{Statements and Declarations}
No potential conflict of interest was reported by the authors.

\setlength\bibsep{0.5pt}
\bibliographystyle{abbrvnat}
\bibliography{main_elsevier}

\newpage
\appendix
\section{}\label{sec:appendix}
\setcounter{figure}{0}

\subsection{Additional Results on \model Interpretability}\label{appendix:additional_results_interpretability}
\begin{figure}[hbt!]
    \centering
        \subfloat[$\mu$ contributions ]{\includegraphics[width=0.5\textwidth]{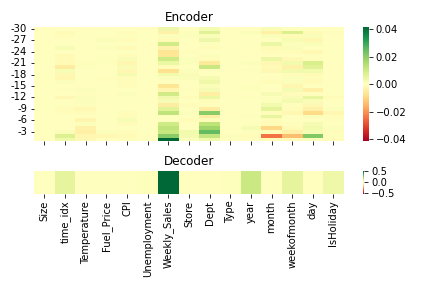}
        \label{fig:walmart_sample0_mu}
        }
        \subfloat[$\sigma$ contributions]{\includegraphics[width=0.5\textwidth]{xai_figs/walmart_mu_sample_0_target_4.png}
        \label{fig:walmart_sample0_sigma}
        }\\
    \caption{Contribution scores obtained from \model for the first sample and 4th target horizon of the Walmart test set. Figure (a) shows the contributions to the $\mu$ and (b) shows contributions to the $\sigma$.}
    \label{fig:walmart_sample_cont}
\end{figure}

\begin{figure}[hbt!]
    \centering
        \subfloat[$\mu$ average importance ]{\includegraphics[width=0.5\textwidth]{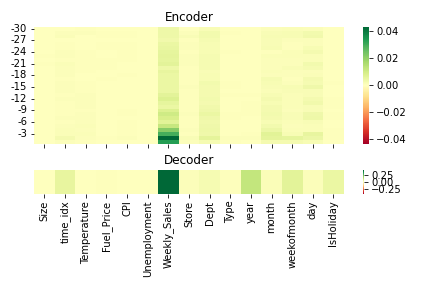}
        \label{fig:walmart_avg_mu}
        }
        \subfloat[$\sigma$ average importance]{\includegraphics[width=0.5\textwidth]{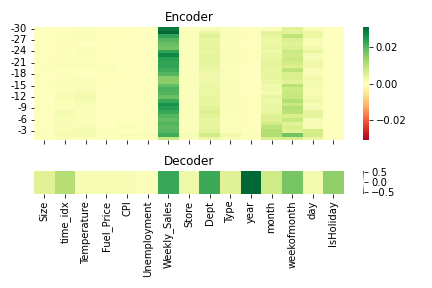}
        \label{fig:walmart_avg_sigma}
        }\\
    \caption{Contribution scores obtained from \model for all samples, and averaged over samples to obtain global importance scores. Figure (a) shows the average importance to the $\mu$ and (b) shows average importance to the $\sigma$ in Walmart dataset.}
    \label{fig:walmart_avg_imp}
\end{figure}

Figure~\ref{fig:walmart_sample_cont} shows the contribution scores obtained from the model for a single sample, and Figure~\ref{fig:walmart_avg_imp} shows the overall importance of the features for the model. 
Overall, we find that the target feature (Weekly Sales) has the most significant impact, particularly at the decoder step, and at the recent timesteps of the encoder. 
Looking closely at the contribution scores, we also observe that various timesteps of ``Dept'', ``year'', ``month'', ''weekofmonth'', and ``day'' features contribute to the predictions. 
However, looking at the importance scores in Figure~\ref{fig:walmart_avg_imp}, we observe that the overall impact of these features is insignificant for the predicted $\mu$, but significant for the predicted $\sigma$.

\begin{figure}[hbt!]
    \centering
        \subfloat[$\mu$ contributions ]{\includegraphics[width=0.5\textwidth]{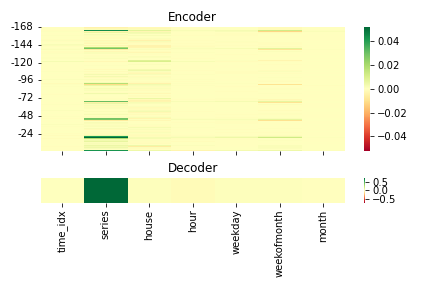}
        \label{fig:electricity_sample0_mu}
        }
        \subfloat[$\sigma$ contributions]{\includegraphics[width=0.5\textwidth]{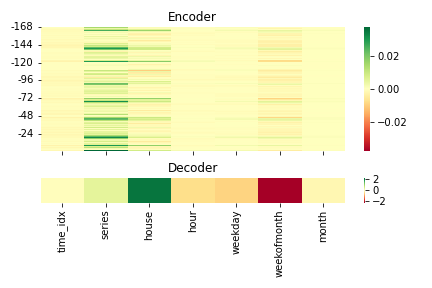}
        \label{fig:electricity_sample0_sigma}
        }\\
    \caption{Contribution scores obtained from \model for the first sample and 4th target horizon of the Electricity test set. Figure (a) shows the contributions to the $\mu$ and (b) shows contributions to the $\sigma$.}
    \label{fig:electricity_sample_cont}
\end{figure}

\begin{figure}[hbt!]
    \centering
        \subfloat[$\mu$ average importance ]{\includegraphics[width=0.5\textwidth]{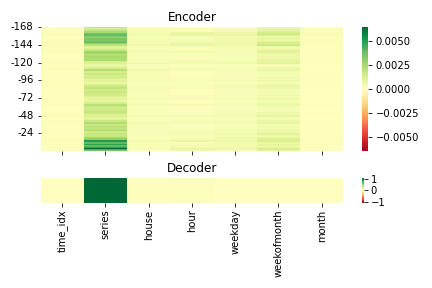}
        \label{fig:electricity_avg_mu}
        }
        \subfloat[$\sigma$ average importance]{\includegraphics[width=0.5\textwidth]{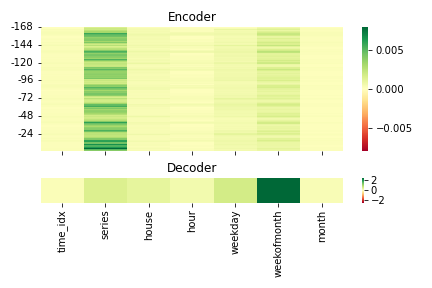}
        \label{fig:electricity_avg_sigma}
        }\\
    \caption{Contribution scores obtained from \model for all samples, and averaged over samples to obtain global importance scores. Figure (a) shows the average importance to the $\mu$ and (b) shows average importance to the $\sigma$ in Electricity dataset.}
    \label{fig:electricity_avg_imp}
\end{figure}

Figure~\ref{fig:electricity_sample_cont} shows the contribution scores obtained from the model for a single sample, and Figure~\ref{fig:electricity_avg_imp} shows the overall importance of the features for the model. 
Similar to the Walmart dataset, we observe that the target feature (series) has the most significant impact, particularly at the decoder timestep. 
Comparing contribution scores to the overall importance scores, we find that in different samples, the model tends to attend different past encoder timesteps of the ``series'' feature. 
This is unlike the Walmart dataset, in which the model attends more to the recent timesteps of the encoder. This suggests that, for the Walmart dataset, the recent timesteps are the most informative, whereas, for the Electricity, the seasonal values of the target feature ($t-24, t-48 \cdots$) are the most informative. 
Looking at the contribution scores for $\sigma$ in Figure~\ref{fig:electricity_sample0_sigma}, we find that the ``house'' feature, which indicates from which house the electricity load information is collected, leads to an increase in the predicted variance. 
On the other hand, decoder timesteps of the time covariates (``hour'', weekday'', and ``weekofmonth'') reduce the predicted variance.

\end{document}